\documentclass{article}
\usepackage{stywhispers}

\usepackage{amssymb}
\usepackage{amsmath}

\usepackage{cite}

\usepackage[nolist,nohyperlinks]{acronym}

\usepackage[inline,shortlabels]{enumitem}

\usepackage{tabularx}
\usepackage{multirow}

\usepackage{siunitx}
\usepackage{tikz}
\usepackage{pgfplots}

\usepackage[hidelinks]{hyperref}

\DeclareSIUnit{\pixel}{pixel}
\DeclareSIUnit{\pixels}{pixels}
\DeclareSIUnit{\band}{bands}
\DeclareSIUnit{\sband}{spectral \ bands}
\DeclareSIUnit{\patch}{patch}
\DeclareSIUnit{\patches}{patches}
\DeclareSIUnit{\tiles}{tiles}
\DeclareSIUnit{\epoch}{epochs}
\DeclareSIUnit{\megapixel}{MP}
\DeclareSIUnit{\bit}{bit}
\DeclareSIUnit{\day}{days}
\DeclareSIUnit{\flops}{FLOPS}
\DeclareSIUnit{\nothing}{\relax}

\usetikzlibrary{arrows}
\usetikzlibrary{positioning}
\usetikzlibrary{calc}
\usetikzlibrary{shapes.geometric}
\usetikzlibrary{patterns}
\usetikzlibrary{patterns.meta}
\usetikzlibrary{spy}
\usetikzlibrary{fit}

\tikzset{
    align=center,
    node distance=1mm and 5mm,
    mark size=2.5pt,
}

\tikzstyle{every node} = [very thick]
\tikzstyle{arrow} = [thick,draw,->,>=stealth]
\tikzstyle{arrow-reverse} = [thick,draw,<-,>=stealth]
\tikzstyle{add} = [circle,draw,inner sep=1.5pt]
\tikzstyle{block} = [rectangle,rounded corners=1pt,draw,minimum width=2cm,minimum height=8mm]
\tikzstyle{conv1} = [trapezium,draw,shape border rotate=90,trapezium angle=78]
\tikzstyle{conv2} = [trapezium,draw,shape border rotate=270,trapezium angle=78]
\tikzstyle{conv3} = [trapezium,draw,shape border rotate=0,trapezium angle=78]
\tikzstyle{conv4} = [trapezium,draw,shape border rotate=180,trapezium angle=78]
\tikzstyle{dim} = [ellipse,draw,densely dashed,minimum width=2cm,minimum height=2mm]
\tikzstyle{act} = [rectangle,rounded corners=1pt,draw,minimum width=5mm,minimum height=20mm]

\tikzstyle{square} = [rectangle,rounded corners=1pt,minimum width=1mm,minimum height=1mm]
\tikzstyle{token} = [rectangle,rounded corners=2mm,minimum width=8mm,minimum height=5mm,draw,anchor=west,fill=magenta]
\tikzstyle{postoken} = [rectangle,rounded corners=2mm,minimum width=6mm,minimum height=1mm,draw,anchor=west,fill=cyan]

\pgfplotsset{
    compat=newest,
    grid=both,
    grid style={line width=.1pt, draw=gray!10},
    major grid style={line width=.2pt,draw=gray!50},
    every axis plot/.append style={line width=0.8pt},
}

\newcommand{\matr}[1]{\mathbf{#1}} 
\newcommand{\vect}[1]{\mathbf{#1}}

\title{HyCoT: A Transformer-Based Autoencoder for Hyperspectral Image Compression}

\name{Martin Hermann Paul Fuchs$^{1}$, Behnood Rasti$^{1,2}$, Begüm Demir$^{1,2}$}
\address{
    $^1$Faculty of Electrical Engineering and Computer Science, Technische Universität Berlin, Germany\\%
    $^2$BIFOLD - Berlin Institute for the Foundations of Learning and Data, Germany
}

\apptocmd{\sloppy}{\hbadness 4000\relax}{}{}

\begin{document}

\begin{acronym}
    \acro{enmap}[EnMAP]{Environmental Mapping and Analysis Program}
    \acro{l2a}[L2A]{Level 2A}

    \acro{rs}[RS]{remote sensing}
    \acro{gpu}[GPU]{graphics processing unit}
    \acro{rgb}[RGB]{red, green and blue}
    \acro{hsi}[HSI]{hyperspectral image}
    \acro{cnn}[CNN]{convolutional neural network}
    \acro{ann}[ANN]{artificial neural network}
    
    \acro{psnr}[PSNR]{peak signal-to-noise ratio}
    \acro{mse}[MSE]{mean squared error}
    \acro{cr}[CR]{compression ratio}

    \acro{decibel}[\si{\decibel}]{decibels}
    \acro{gsd}[GSD]{ground sample distance}

    \acro{flops}[FLOPs]{floating point operations}

    \acro{cae}[CAE]{convolutional autoencoder}

    \acro{leakyrelu}[LeakyReLU]{leaky rectified linear unit}
    \acro{prelu}[PReLU]{parametric rectified linear unit}
    
    \acro{a1dcae}[A1D-CAE]{Adaptive 1-D Convolutional Autoencoder}
    \acro{1dcae}[1D-CAE]{1D-Convolutional Autoencoder}
    \acro{sscnet}[SSCNet]{Spectral Signals Compressor Network}
    \acro{3dcae}[3D-CAE]{3D Convolutional Auto-Encoder}

    \acro{ct}[CT]{compression token}
    \acro{mlp}[MLP]{multilayer perceptron}

    \acro{1d}[1D]{one-dimensional}
    \acro{2d}[2D]{two-dimensional}
    \acro{3d}[3D]{three-dimensional}

    \acro{sa}[SA]{self-attention}
    \acro{msa}[MSA]{multi-head self-attention}
    \acro{ln}[LN]{layer normalization}

    \acro{ourmodel}[HyCoT]{Hyperspectral Compression Transformer}
\end{acronym}

\maketitle

\begin{abstract}
The development of learning-based \ac{hsi} compression models has recently attracted significant interest. Existing models predominantly utilize convolutional filters, which capture only local dependencies.
Furthermore, they often incur high training costs and exhibit substantial computational complexity.
To address these limitations, in this paper we propose \ac{ourmodel} that is a transformer-based autoencoder for pixelwise \ac{hsi} compression. 
Additionally, we apply a simple yet effective training set reduction approach to accelerate the training process.
Experimental results on the HySpecNet-11k dataset demonstrate that \ac{ourmodel} surpasses the state of the art across various \aclp{cr} by over \SI{1}{\decibel} of \acs{psnr} with significantly reduced computational requirements.
Our code and pre-trained weights are publicly available at \url{https://git.tu-berlin.de/rsim/hycot}.
\end{abstract}


\section{Introduction}
\label{sec:introduction}
Hyperspectral sensors acquire images with high spectral resolution, producing data across hundreds of observation channels. This extensive spectral information allows for detailed analysis and differentiation of materials based on their spectral signatures.
However, the large number of spectral bands in \acfp{hsi} presents a significant challenge, namely the vast amount of data generated by hyperspectral sensors.
While the volume of hyperspectral data archives is growing rapidly, transmission bandwidth and storage space are expensive and limited.
To address this problem, \ac{hsi} compression methods have been developed to efficiently transmit and store hyperspectral data.

Generally, there are two categories of \ac{hsi} compression:
\begin{enumerate*}[i)]
    \item traditional methods \cite{lim2001compression, penna2006new, du2007hyperspectral}; and
    \item learning-based methods \cite{kuester20211d,la2022hyperspectral,chong2021end, sprengel2024learning, guo2023hyperspectral}.
\end{enumerate*}
Most traditional methods are based on transform coding in combination with a quantization step and entropy coding.
In contrast, learning-based methods train an \ac{ann} to extract representative features and reduce the dimensionality of the latent space via consecutive downsampling operations.

Recent studies indicate that learning-based \ac{hsi} compression methods can preserve reconstruction quality at higher \acp{cr} compared to traditional compression methods \cite{altamimi2021systematic}.
Thus, the development of learning-based \ac{hsi} compression methods has attracted great attention.
The current state of the art in this field is formed by \acp{cae} \cite{kuester20211d, la2022hyperspectral, chong2021end}.
The \ac{1dcae} model presented in \cite{kuester20211d} compresses the spectral content without considering spatial redundancies by stacking multiple blocks of \acs{1d} convolutions, pooling layers and \ac{leakyrelu} activations. Although high-quality reconstructions can be achieved with this approach, the pooling layers limit the achievable \acp{cr} to $2^n$. Another limitation of this approach is the increasing computational complexity with higher \acp{cr} due to the deeper network architecture.
Unlike \cite{kuester20211d}, to achieve spatial compression, \ac{sscnet} that focuses on the spatial redundancies via \acs{2d} convolutional filters is introduced in \cite{la2022hyperspectral}. In \ac{sscnet}, compression is achieved by \acs{2d} max pooling, while the final \ac{cr} is adapted via the latent channels inside the bottleneck. With \ac{sscnet} much higher \acp{cr} can be achieved at the cost of spatially blurry reconstructions.
To achieve joint compression of spectral and spatial dependencies, \ac{3dcae} is introduced in \cite{chong2021end}. \ac{3dcae} applies spatio-spectral compression via \acs{3d} convolutional filter kernels. Residual blocks allow gradients to flow through the network directly and improve the model performance. 
In general, \acp{cae} fail to capture long-range spatial and spectral dependencies and their stacked convolutional layers make the architectures computationally expensive. 

It is important to note that most of the deep learning models require large training sets to effectively optimize their parameters during training.
One the one hand, training these models using large training sets allows them to generalize well. 
On the other hand, training often requires hundreds of epochs for model weight convergence \cite{fuchs2023hyspecnet}, which results in high training costs. While training time can be reduced through parallelization on multiple \acsp{gpu}, energy consumption and hardware requirements are still limiting factors.

To address these problems, we propose the \acf{ourmodel} model that exploits a transformer-based autoencoder for \ac{hsi} compression. \ac{ourmodel} leverages long-range spectral dependencies for latent space encoding. For fast reconstruction, \ac{ourmodel} employs a lightweight decoder. According to our knowledge, this is the first time a transformer-based model is investigated for \ac{hsi} compression. To accelerate training while preserving reconstruction quality, we apply random sampling for reducing the training set.
Experimental results demonstrate that \ac{ourmodel} achieves higher reconstruction quality across multiple \acp{cr}, while reducing both training costs and computational complexity compared to state-of-the-art methods.

\begin{figure*}
    \centering
    \begin{tikzpicture}
        \node (in) {\includegraphics[width=15mm]{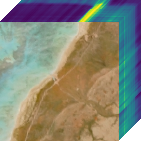}};

        \node[right=of in] (spec) {
            \centering
            \begin{tikzpicture}[scale=0.1]
                \draw[color=blue,solid]  plot[smooth,tension=.7] coordinates{
                    (0,3) (2,0) (4,4) (6,2) (8,1) (10,4)
                };
                \draw[color=red,solid]  plot[smooth,tension=.7] coordinates{(10,4) (11,4)};
            \end{tikzpicture}
        };

        \node[above=of spec] (spec1) {
            \centering
            \begin{tikzpicture}[scale=0.1]
                \draw[color=blue,solid]  plot[smooth,tension=.7] coordinates{
                    (0,0) (2,3) (4,2) (6,4) (8,1) (10,2)
                };
                \draw[color=red,solid]  plot[smooth,tension=.7] coordinates{(10,2) (11,2)};
            \end{tikzpicture}
        };

        \node[above=of spec1] (spec2) {
            \centering
            \begin{tikzpicture}[scale=0.1]
                \draw[color=blue,solid]  plot[smooth,tension=.7] coordinates{
                    (0,2) (2,3) (4,2) (6,4) (8,1) (10,0)
                };
                \draw[color=red,solid]  plot[smooth,tension=.7] coordinates{(10,0) (11,0)};
            \end{tikzpicture}
        };

        \node[below=of spec] (spec3) {\rotatebox{90}{$\cdots$}};

        \node[below=of spec3] (spec4) {
            \centering
            \begin{tikzpicture}[scale=0.1]
                \draw[color=blue,solid]  plot[smooth,tension=.7] coordinates{
                    (0,1) (2,4) (4,2) (6,3) (8,1) (10,2)
                };
                \draw[color=red,solid]  plot[smooth,tension=.7] coordinates{(10,2) (11,2)};
            \end{tikzpicture}
        };

        \node[right=of spec] (pixel) {\rotatebox{90}{$\cdots$}};

        \draw[decorate,decoration={brace,amplitude=4mm,mirror,raise=4mm}] ($(pixel)+(0,2.2)$) -- ($(pixel)+(0,-2.2)$);

        \foreach \y in {20,17,14,10.5,7.5,4.5,-4.5,-7.5,-10.5,-14} {
            \node[square,right=of spec,yshift=\y mm,fill=blue] {};
        };
        \foreach \y in {-17,-20} {
            \node[square,right=of spec,yshift=\y mm,fill=red] {};
        };

        \node[right=of pixel,block,minimum height=42mm,minimum width=5mm,fill=pink] (patchembed) {\rotatebox{90}{Linear Projection}};

        \node[token,draw=white,fill=white,right=of patchembed] (tokens) {\rotatebox{90}{$\cdots$}};
        \node[token,fill=brown] at ($(patchembed.east)+(0.5,2.65)$) (compressiontoken) {\acs{ct}};
        \node[token] at ($(patchembed.east)+(0.5,1.7)$) (token1) {};
        \node[token] at ($(patchembed.east)+(0.5,0.75)$) (token2) {};
        \node[token] at ($(patchembed.east)+(0.5,-0.75)$) (token3) {};
        \node[token] at ($(patchembed.east)+(0.5,-1.7)$) (token4) {};

        \node[right=of tokens] (adds) {\rotatebox{90}{$\cdots$}};
        \node[add,right=of compressiontoken] (add0) {+};
        \node[add,right=of token1] (add1) {+};
        \node[add,right=of token2] (add2) {+};
        \node[add,right=of token3] (add3) {+};
        \node[add,right=of token4] (add4) {+};

        \node[postoken,below=of add0,yshift=1mm,xshift=-5mm] (pos0) {\footnotesize{0}};
        \node[postoken,below=of add1,yshift=1mm,xshift=-5mm] (pos1) {\footnotesize{1}};
        \node[postoken,below=of add2,yshift=1mm,xshift=-5mm] (pos2) {\footnotesize{2}};
        \node[postoken,below=of add3,yshift=1mm,xshift=-5mm] (pos3) {\footnotesize{n-1}};
        \node[postoken,below=of add4,yshift=1mm,xshift=-5mm] (pos4) {\footnotesize{n}};

        \node[block,right=of adds,minimum height=50mm,yshift=4mm,minimum width=5mm,fill=green] (trafo) {\rotatebox{90}{Transformer Encoder}};

        \node[token,fill=brown] at ($(trafo.east)+(0.5,2.25)$) (ct) {\acs{ct}};

        \node[dim,right=of trafo,yshift=-4mm,xshift=-0.53581pt] (lat) {Latent\\Space};

        \node[conv4,above=of lat,fill=lightgray,yshift=3mm] (mlpenc) {\acs{mlp}\\Encoder};

        \node[conv1,right=of lat,fill=lightgray] (mlpdec) {\acs{mlp}\\Decoder};

        \draw[decorate,decoration={brace,amplitude=4mm,mirror,raise=-13mm}] ($(mlpdec)+(0,2.2)$) -- ($(mlpdec)+(0,-2.2)$);

        \node[right=of mlpdec] (recspec) {
            \centering
            \begin{tikzpicture}[scale=0.1]
                \draw[color=blue,solid]  plot[smooth,tension=.7] coordinates{
                    (0,3) (2,0) (4,4) (6,2) (8,1) (10,4)
                };
                \draw[color=red,solid]  plot[smooth,tension=.7] coordinates{(10,4) (11,4)};
            \end{tikzpicture}
        };

        \node[above=of recspec] (recspec1) {
            \centering
            \begin{tikzpicture}[scale=0.1]
                \draw[color=blue,solid]  plot[smooth,tension=.7] coordinates{
                    (0,0) (2,3) (4,2) (6,4) (8,1) (10,2)
                };
                \draw[color=red,solid]  plot[smooth,tension=.7] coordinates{(10,2) (11,2)};
            \end{tikzpicture}
        };

        \node[above=of recspec1] (recspec2) {
            \centering
            \begin{tikzpicture}[scale=0.1]
                \draw[color=blue,solid]  plot[smooth,tension=.7] coordinates{
                    (0,2) (2,3) (4,2) (6,4) (8,1) (10,0)
                };
                \draw[color=red,solid]  plot[smooth,tension=.7] coordinates{(10,0) (11,0)};
            \end{tikzpicture}
        };

        \node[below=of recspec] (recspec3) {\rotatebox{90}{$\cdots$}};

        \node[below=of recspec3] (recspec4) {
            \centering
            \begin{tikzpicture}[scale=0.1]
                \draw[color=blue,solid]  plot[smooth,tension=.7] coordinates{
                    (0,1) (2,4) (4,2) (6,3) (8,1) (10,2)
                };
                \draw[color=red,solid]  plot[smooth,tension=.7] coordinates{(10,2) (11,2)};
            \end{tikzpicture}
        };

        \node[right=of recspec] (out) {\includegraphics[width=15mm]{img/hsi.pdf}};

        \draw[arrow,blue] ($(in)+(-0.5,0.45)$) -- (spec.west);
        \draw[arrow,blue] ($(in)+(-0.6,0.45)$)  -- (spec1.west);
        \draw[arrow,blue] ($(in)+(-0.7,0.45)$) -- (spec2.west);
        \draw[arrow,blue] ($(in)+(0.4,-0.7)$)  -- (spec4.west);

        \draw[arrow] ($(pixel)+(0.1,1.7)$) -- ++(0.6,0);
        \draw[arrow] ($(pixel)+(0.1,0.75)$) -- ++(0.6,0);
        \draw[arrow] ($(pixel)+(0.1,-0.75)$) -- ++(0.6,0);
        \draw[arrow] ($(pixel)+(0.1,-1.7)$) -- ++(0.6,0);

        \draw[arrow] ($(patchembed.east)+(0,1.7)$) -- ++(0.5,0);
        \draw[arrow] ($(patchembed.east)+(0,0.75)$) -- ++(0.5,0);
        \draw[arrow] ($(patchembed.east)+(0,-0.75)$) -- ++(0.5,0);
        \draw[arrow] ($(patchembed.east)+(0,-1.7)$) -- ++(0.5,0);

        \draw[arrow] (compressiontoken.east) -- (add0);
        \draw[arrow] (token1.east) -- (add1);
        \draw[arrow] (token2.east) -- (add2);
        \draw[arrow] (token3.east) -- (add3);
        \draw[arrow] (token4.east) -- (add4);

        \draw[arrow] (add0.east) -- ++(0.5,0);
        \draw[arrow] (add1.east) -- ++(0.5,0);
        \draw[arrow] (add2.east) -- ++(0.5,0);
        \draw[arrow] (add3.east) -- ++(0.5,0);
        \draw[arrow] (add4.east) -- ++(0.5,0);

        \draw[arrow] (pos0.east) -| (add0.south);
        \draw[arrow] (pos1.east) -| (add1.south);
        \draw[arrow] (pos2.east) -| (add2.south);
        \draw[arrow] (pos3.east) -| (add3.south);
        \draw[arrow] (pos4.east) -| (add4.south);

        \draw[arrow] ($(trafo.east)+(0,2.25)$) -- (ct);

        \draw[arrow] (ct) -| (mlpenc);

        \draw[arrow] (mlpenc) -- (lat);
        \draw[arrow] (lat) -- (mlpdec);

        \draw[arrow,blue] (recspec.east) -- ($(out)+(-0.5,0.45)$);
        \draw[arrow,blue] (recspec1.east) -- ($(out)+(-0.6,0.45)$);
        \draw[arrow,blue] (recspec2.east) --($(out)+(-0.7,0.45)$);
        \draw[arrow,blue] (recspec4.east) -- ($(out)+(0.4,-0.7)$);

        \draw[dashed] ($(pixel)+(-0.25,2.15)$) rectangle ($(pixel)+(0.1,1.25)$);
        \draw[dashed] ($(pixel)+(-0.25,1.2)$) rectangle ($(pixel)+(0.1,0.3)$);
        \draw[dashed] ($(pixel)+(-0.25,-0.3)$) rectangle ($(pixel)+(0.1,-1.2)$);
        \draw[dashed] ($(pixel)+(-0.25,-1.25)$) rectangle ($(pixel)+(0.1,-2.15)$);
    \end{tikzpicture}
    \caption{Overview of our proposed \ac{ourmodel} model. For each pixel of a given \ac{hsi}, the spectral signature is padded. Neighbouring spectral bands are grouped and embedded by a linear projection \cite{hong2021spectralformer} to form the transformer tokens. A \ac{ct} is prepended and a learned position embedding is added. The sequence of tokens is fed into the transformer encoder \cite{vaswani2017attention}. Then, the \ac{ct} is extracted and projected in the \ac{mlp} encoder to fit the target \ac{cr}. An \ac{mlp} decoder is applied to reconstruct the full spectral signature. Finally, the decoded pixels are reassembled to form the reconstructed \ac{hsi}.}
    \label{fig:methodology-block_diagram}
\end{figure*}

\section{Methodology}
\label{sec:methodology}
Let $\matr{X} \in \mathbb{R}^{H \times W \times C}$ be an \ac{hsi} with image height $H$, image width $W$ and $C$ spectral bands.
The objective of lossy \ac{hsi} compression is to encode $\matr{X}$ into a decorrelated latent space $\matr{Y} \in \mathbb{R}^{\Sigma \times \Omega \times \Gamma}$ that represents $\matr{X}$ with minimal distortion $d: \matr{X} \times \matr{\hat{X}} \rightarrow [0, \infty)$ after reconstructing $\matr{\hat{X}} \in \mathbb{R}^{H \times W \times C}$ on the decoder side.

To efficiently and effectively compress $\matr{X}$, we present \ac{ourmodel}.
\ac{ourmodel} is a learning-based \ac{hsi} compression model that employs the SpectralFormer \cite{hong2021spectralformer} as a feature extractor on the encoder side, and a lightweight \acf{mlp} for lossy reconstruction on the decoder side.
Our model is illustrated in \autoref{fig:methodology-block_diagram}.
In the following subsections, we provide a comprehensive explanation of the proposed model.
\subsection{\acs{ourmodel} Encoder}
The \ac{ourmodel} encoder $E_\Phi: \matr{X} \rightarrow \matr{Y}$ uses the SpectralFormer \cite{hong2021spectralformer} as a backbone for spectral feature extraction. It aggregates long-range spectral dependencies via transformer blocks \cite{vaswani2017attention} inside a \acf{ct} and then projects the \ac{ct} into the latent space using an \ac{mlp} to fit the desired target \ac{cr}.

In detail, each pixel $\vect{x}$ of $\matr{X}$ is processed separately.
First, spectral groups $\vect{g}^i$ are formed by nonoverlapping grouping of $g_d$ neighbouring spectral bands, where $g_d$ denotes the group depth.
To make the number of bands divisible by $g_d$, $\vect{x}$ is padded from $C$ to $C_\text{pad} = C + \Delta_\text{pad}$, forming $\vect{x}_\text{pad}$, where $\Delta_\text{pad} = g_d - C \bmod g_d$.
The spectral groups are linearly projected to the embedding dimension $d_\text{emb}$ using a learnable matrix $\matr{P}$. The resulting embedding vectors (so-called tokens) are defined as $\vect{\tilde{t}}^i = \matr{P} \vect{g}^i$.
We propose prepending a learnable \ac{ct} $\vect{\tilde{t}}^0$, which is also encoded by the transformer encoder to capture the spectral information of the pixel.
The overall number $n_t$ of tokens per pixel thus equals $n_t = C_\text{pad} / g_d + 1$.
Afterwards, a learned position embedding $\vect{l}^i$ is added to the patch embeddings to retain the positional information of the spectral groups as $\vect{t}^i = \vect{\tilde{t}}^i + \vect{l}^i$.

The sequence of tokens is then fed into the transformer encoder that stacks $L$ transformer blocks.
At each transformer block $l = 1, \ldots, L$, \ac{msa} is applied followed by an \ac{mlp} \cite{vaswani2017attention}.
For both, a residual connection \cite{he2016deep} is employed, followed by a \ac{ln} \cite{ba2016layer}:
\begin{align}
    \vect{t}'_l &= \acs{msa} \left( \acs{ln} \left( \vect{t}_{l-1} \right) \right) + \vect{t}_{l-1} ,
    \label{eq:msa} \\
    \vect{t}_l &= \acs{mlp} \left( \acs{ln} \left( \vect{t}'_l \right) \right) + \vect{t}'_l ,
    \label{eq:mlp}
\end{align}
where \ac{msa} is an extension of \ac{sa} that runs $k$ self-attention operations (heads) in parallel and projects their concatenated outputs \cite{vaswani2017attention}:
\begin{align}
    \ac{msa} \left( \vect{t} \right) = \left[ \ac{sa}_1 \left( \vect{t} \right); \ac{sa}_2 \left( \vect{t} \right); \ldots; \ac{sa}_k \left( \vect{t} \right) \right] \matr{U}_\text{MSA}.
\end{align}
\ac{sa} computes a weighted sum over all values $\vect{v}$ for each token $\vect{t}$ and the attention weights $\matr{A}_{ij}$ are based on the pairwise similarity between the query $\vect{q}^i$ and key $\vect{k}^j$ of two respective tokens \cite{vaswani2017attention}:
\begin{align}
    \left[ \vect{q}, \vect{k}, \vect{v} \right] &= \vect{t} \matr{U}_\text{qkv}, \\
    \matr{A} &= \text{softmax} \left( \frac{\vect{q} \vect{k}^T}{\sqrt{d_\text{emb} / k}} \right), \\
    \text{SA} \left( \vect{t} \right) &= \matr{A} \vect{v} .
\end{align}

The \ac{ct} is extracted and sent to the \ac{mlp} encoder that adapts the latent space channels $\Gamma$ to fit the target \ac{cr}. 
\ac{ourmodel} compresses only the spectral content, and therefore the \ac{cr} can be simplified as follows:
\begin{align}
    \ac{cr} = \frac{C}{\Gamma} .
    \label{eq:cr}
\end{align}
The \ac{mlp} encoder initially applies a linear transformation of the \ac{ct} with the embedding dimension $d_\text{emb}$ to a hidden dimension $d_\text{hid}$, followed by a \ac{leakyrelu} to introduce nonlinearity.
Another linear transformation then adapts the \ac{cr} by having $\Gamma$ output features.
Finally, the sigmoid activation function is applied to rescale the latent space.

\subsection{\acs{ourmodel} Decoder}
The latent space $\matr{Y}$ only consists of a few latent channels $\Gamma$ per pixel.
Consequently, a simple feedforward \ac{ann} with nonlinear activation functions is sufficient for a qualitative reconstruction.
This enables fast decoding for real-time applications or frequent access to data archives with low latency.
Thus, similar to the last block on the encoder side, the \ac{ourmodel} decoder $D_\theta: \matr{Y} \rightarrow \matr{\hat{X}}$ is lightweight and consists of only a simple \ac{mlp} with one hidden dimension $d_\text{hid}$ that upsamples the latent channel dimension $\Gamma$ back to \ac{hsi} input bands $C$.
\ac{leakyrelu} is used after the initial layer and the sigmoid function after the hidden layer to scale the data back into the range \SIrange{0}{1}{} for the decoder output.
Finally, the decoded pixels are reassembled to form the reconstructed image $\matr{\hat{X}}$.

\section{Experimental Results}
\label{sec:experiments}

The experiments were conducted on HySpecNet-11k \cite{fuchs2023hyspecnet} that is a large-scale hyperspectral benchmark dataset based on \num{250} tiles acquired by the \ac{enmap} satellite \cite{guanter2015enmap}. HySpecNet-11k includes \num{11483} nonoverlapping \acp{hsi}, each of them having \qtyproduct{128x128}{} \si{\pixels} and \SI{202}{\band} with a \acl{gsd} of \SI{30}{\meter}.
Our code is implemented in PyTorch based on the CompressAI \cite{begaint2020compressai} framework. 
Training runs were carried out on a single NVIDIA A100 SXM4 80 GB \acs{gpu} using the Adam optimizer \cite{kingma2014adam}.
For the \ac{ourmodel} transformer encoder, we used $d_\text{emb} = \num{64}$, $L = \num{5}$ and $k = \num{4}$ as suggested in \cite{hong2021spectralformer}.
We empirically fixed the group depth $g_d = \num{4}$ and the hidden \ac{mlp} dimension of both encoder and decoder to $d_\text{hid} = \num{1024}$.

\subsection{Training Set Reduction}
\label{subsec:results-reduction}
Spectral information of pixels of the same material can be highly similar, leading to redundant information for the training stage of learning-based models.
We aim to reduce the number of training samples by considering only a subset of the samples.
To this end, we exploit random sampling within the training \acp{hsi} for efficiently reducing the training time.
Instead of using all $W \times H$ \si{\pixels} of each training \ac{hsi} $\matr{X} \in \mathbb{R}^{H \times W \times C}$, we only consider a fraction of that by randomly sampling $n$ \si{\pixels} of each \ac{hsi} in each training epoch with $n \ll W \cdot H$. As a quantitative measure, we define the training set reduction factor $r = \frac{W \cdot H}{n}.$
In this subsection, we analyse the efficiency of our suggested training set reduction approach.
To this end, we fix $r$ to have an optimal trade-off between training costs and reconstruction quality.
In contrast to the training set, the test set is left unchanged to achieve comparable results.

To reduce the computational costs associated with training \ac{ourmodel} from scratch for each $r$, we employed a new HySpecNet-11k split, denoted as the mini split.
The mini split represents a condensed version of HySpecNet-11k, which we recommend to use for hyperparameter tuning.
To obtain the mini split, we followed \cite{fuchs2023hyspecnet} and  divided the dataset randomly into 
\begin{enumerate*}[i)]
    \item a training set that includes \SI{70}{\percent} of the \acp{hsi},
    \item a validation set that includes \SI{20}{\percent} of the \acp{hsi}, and
    \item a test set that includes \SI{10}{\percent} of the \acp{hsi}.
\end{enumerate*}
However, the mini split only uses the center \ac{hsi} of each tile used to create HySpecNet-11k. This selection strategy reduces the dataset size to \num{250} \acp{hsi}, while maintaining the global diversity of HySpecNet-11k (\num{11483} \acp{hsi}).

\autoref{tab:img-sizes} shows the results of training \ac{ourmodel} for the highest considered $\ac{cr} = \num[round-mode=places,round-precision=2]{28.857}$ on the HySpecNet-11k mini split using different $r$.
For this experiment, the \ac{ourmodel} models were trained for \SI{1000}{\epoch} with a learning rate of \num{1e-4}.
The table shows that reconstruction quality can be preserved when training with a randomly subsampled training set.
In detail, reconstruction quality for $r = \num{64}$ is slightly higher than considering the complete training set ($r = \num{1}$).
This indicates high redundancy among the pixels within the training set, suggesting that not all pixels are necessary to achieve robust generalization performance.
We would like to note that even higher values of $r$ can be considered for even further training cost reduction at a slightly reduced reconstruction quality.

\begin{table}
    \centering
    \caption{\acs{ourmodel} trained on HySpecNet-11k (mini split) with different values of the training set reduction factor $r$ for a fixed \ac{cr} of \num[round-mode=places,round-precision=2]{28.857}. Reconstruction quality is evaluated as \ac{psnr} on the test set.} 
    \label{tab:img-sizes}
    \begin{tabular}{cc}
        \hline
        \multicolumn{1}{c}{$r$} & \multicolumn{1}{c}{\acs{psnr} [\si{\decibel}]} \\
        \hline
        \num{1}&\num{45.612} \\
        \num{4} & \num{45.550} \\
        \num{16} & \num{45.344} \\
        \num{64} & {\boldmath$\num{45.669}$} \\
        \num{256} & \num{45.193} \\
        \num{1024} & \num{44.886} \\
        \num{4096} & \num{44.509} \\
        \num{16384} & \num{43.977} \\
        \hline
    \end{tabular}
\end{table}

\subsection{Rate-Distortion Evaluation}
\label{subsec:results-ratedistortion}
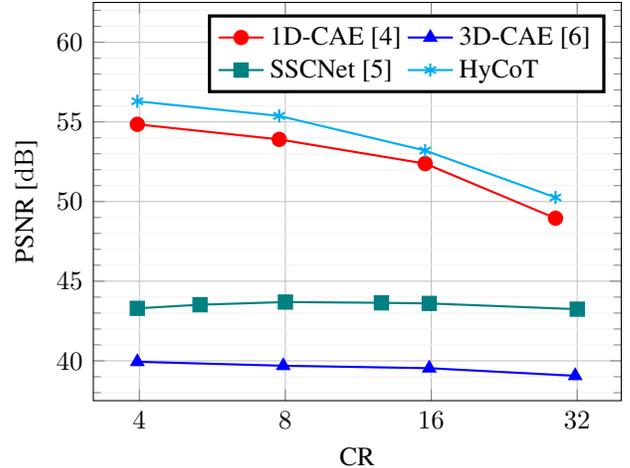
\begin{figure}
    \centering
    \begin{tikzpicture}
        \begin{axis}[
                width=\linewidth,
                height=0.8\linewidth,
                xlabel={\acs{cr}},
                ylabel={\acs{psnr} [\si{\decibel}]},
                legend columns=2,
                transpose legend,
                legend cell align={left},
                legend pos=north east,
                xmode=log,
                log basis x={2},
                log ticks with fixed point,
                minor tick num=4,
                ymin=37.5,
                ymax=62.5,
            ]
            \addplot[red,mark=*] coordinates {
                (3.960784314, 54.84735990895165)        
                (7.769230769, 53.902654146940)          
                (15.538461538, 52.38174220872962)       
                (28.857142857, 48.952538482825375)      
            };
            \addlegendentry{\acs{1dcae} \cite{kuester20211d}};

            \addplot[teal,mark=square*] coordinates {
                (3.96078431373, 43.29476347234514)      
                (8, 43.69138220945994)                  
                (15.8431372549, 43.60347966353098)      
                (32, 43.2428080505795)                  
            };
            \addlegendentry{\acs{sscnet} \cite{la2022hyperspectral}};

            \addplot[blue,mark=triangle*] coordinates {
                (3.96078431373, 39.94142961502075)      
                (7.92156862745, 39.6920463376575)       
                (15.8431372549, 39.5384429163403)       
                (31.6862745098, 39.06108974227706)      
            };
            \addlegendentry{\acs{3dcae} \cite{chong2021end}};

            \addplot[cyan,mark=asterisk] coordinates {
                (3.9608, 56.294)                        
                (7.7692, 55.377)                        
                (15.538, 53.202)                        
                (28.857, 50.257)                        
            };
            \addlegendentry{\acs{ourmodel}};

        \end{axis}
    \end{tikzpicture}
    \caption{Rate-distortion performance on the test set of HySpecNet-11k (easy split).} 
    \label{fig:rdplot}
\end{figure}

In this subsection, we analyse the rate-distortion curves of learning-based \ac{hsi} compression models.
Our experimental results are shown in \autoref{fig:rdplot}.
To achieve these results, we used the HySpecNet-11k easy split as proposed in \cite{fuchs2023hyspecnet}.
We used \acf{mse} as a loss function until convergence on the validation set, which was achieved after \SI{2000}{\epoch} for each experiment using a learning rate of \num{1e-3}.
We trained \ac{ourmodel}, targeting the four specific \acp{cr} of \ac{1dcae} \cite{kuester20211d}, to make our results comparable to the baseline.
We would like to note that \ac{ourmodel} is more versatile in terms of \ac{cr}, see (\ref{eq:cr}), than \acp{cae} since \ac{ourmodel} is not based on strided convolutions or pooling operations.
For the \ac{ourmodel} results, we significantly reduced the training costs by a factor of $r = \num{64}$. 
Yet, \ac{ourmodel} is able to outperform all \ac{cae} baselines at every \ac{cr}, which were trained on the full training set.
Quantitatively, \ac{ourmodel} is at least \SI{1}{\decibel} better than \ac{1dcae} for each \ac{cr} while for $\ac{cr} \approx \num{4}$ outperforming \ac{sscnet} and \ac{3dcae} with \SI{13.00}{\decibel} and \SI{16.36}{\decibel}, respectively.
This shows that \acl{1d} spectral dimensionality reduction is beneficial for \ac{hsi} compression and that long-range spectral dependencies, which are captured inside the transformer encoder of \ac{ourmodel}, are relevant for high quality reconstruction.

\subsection{Computational Complexity Analysis}
In this subsection, we compare the computational complexity of \ac{ourmodel} with those of other models.
In \autoref{tab:compute} we report a comparison of the \acf{flops} and the number of model parameters for multiple \acp{cr}.
From the table, one can see that for low \acp{cr}, \ac{1dcae} \cite{kuester20211d} has few parameters.
However, by increasing \ac{cr}, the parameters rise exponentially, since the network stacks up layers with each added downsampling operation.
The high \ac{flops} of \ac{1dcae} are due to the pixelwise convolutional processing.
In contrast, the computational complexity of \ac{ourmodel} decreases by increasing \acp{cr} since the transformer encoder backbone stays the same and only the output channels of the \ac{mlp} encoder are slimmed, when compressing to a smaller latent space.
Thus, \ac{ourmodel} results in a faster runtime. In addition, increasing the \ac{cr} reduces the number of parameters compared to \ac{1dcae}, where the opposite is the case.
It is worth mentioning that the lightweight decoder of \ac{ourmodel} is suitable for real-time \ac{hsi} reconstruction.
We also report the computational complexity for \ac{sscnet} and \ac{3dcae} in the table. 
\ac{sscnet} and \ac{3dcae} are associated to higher parameter counts and \ac{flops} due to their multi-dimensional convolutional layers and therefore cannot compete with our \ac{ourmodel} model.

\begin{table}
    \centering
    \caption{Computational complexity analysis. \ac{flops} are calculated using a HySpecNet-11k \ac{hsi} of size \qtyproduct{128x128}{} \si{\pixels}.}
    \label{tab:compute}
    \begin{tabular}{ccrr}
        \hline
        Model & \acs{cr} & \multicolumn{1}{c}{\acs{flops}} & \multicolumn{1}{c}{Parameters} \\ 
        \hline
        
        \acs{1dcae} \cite{kuester20211d} & \num[round-mode=places,round-precision=2]{3.9607843137254903} & \num[round-mode=places,round-precision=2]{176.64} \si{\giga\nothing} & {\boldmath$\num{56130}$} \\
        \acs{sscnet} \cite{la2022hyperspectral} & \num[round-mode=places,round-precision=2]{4} & \num[round-mode=places,round-precision=2]{84.18} \si{\giga\nothing} & \num{34136586} \\ 
        \acs{3dcae} \cite{chong2021end} & \num[round-mode=places,round-precision=2]{3.9607843137254903} & \num[round-mode=places,round-precision=2]{121.4} \si{\giga\nothing} & \num{933539} \\
        \acs{ourmodel} & \num[round-mode=places,round-precision=2]{3.9607843137254903} & {\boldmath$\num[round-mode=places,round-precision=2]{6.82}$ \si{\giga\nothing}} & \num{488225} \\ 
        \hline
        
        \acs{1dcae} \cite{kuester20211d} & \num[round-mode=places,round-precision=2]{7.7692} & \num[round-mode=places,round-precision=2]{719.94} \si{\giga\nothing} & {\boldmath$\num{238018}$} \\
        \acs{sscnet} \cite{la2022hyperspectral} & \num[round-mode=places,round-precision=2]{8} & \num[round-mode=places,round-precision=2]{76.56} \si{\giga\nothing} & \num{19240298} \\ 
        \acs{3dcae} \cite{chong2021end} & \num[round-mode=places,round-precision=2]{7.921568627450981} & \num[round-mode=places,round-precision=2]{104.68} \si{\giga\nothing} & \num{677443} \\
        \acs{ourmodel} & \num[round-mode=places,round-precision=2]{7.7692} & {\boldmath$\num[round-mode=places,round-precision=2]{5.98}$ \si{\giga\nothing}}& \num{437000} \\ 
        \hline
        
        \acs{1dcae} \cite{kuester20211d} & \num[round-mode=places,round-precision=2]{15.538461538} & \num[round-mode=places,round-precision=2]{2840} \si{\giga\nothing} & \num{962242} \\
        \acs{sscnet} \cite{la2022hyperspectral} & \num[round-mode=places,round-precision=2]{16} & \num[round-mode=places,round-precision=2]{72.74} \si{\giga\nothing} & \num{11792154} \\
        \acs{3dcae} \cite{chong2021end} & \num[round-mode=places,round-precision=2]{15.843137254901961} & \num[round-mode=places,round-precision=2]{96.32} \si{\giga\nothing} & \num{549395} \\
        \acs{ourmodel} & \num[round-mode=places,round-precision=2]{15.538461538} & {\boldmath$\num[round-mode=places,round-precision=2]{5.54}$ \si{\giga\nothing}} & {\boldmath$\num{410363}$} \\ 
        \hline
        
        \acs{1dcae} \cite{kuester20211d} & \num[round-mode=places,round-precision=2]{28.857} & \num[round-mode=places,round-precision=2]{12180} \si{\giga\nothing} & \num{3852482} \\
        \acs{sscnet} \cite{la2022hyperspectral} & \num[round-mode=places,round-precision=2]{32} & \num[round-mode=places,round-precision=2]{70.84} \si{\giga\nothing} & \num{8068082} \\ 
        \acs{3dcae} \cite{chong2021end} & \num[round-mode=places,round-precision=2]{31.686274509803923} & \num[round-mode=places,round-precision=2]{92.16} \si{\giga\nothing} & \num{485371} \\
        \acs{ourmodel} & \num[round-mode=places,round-precision=2]{28.857} & {\boldmath$\num[round-mode=places,round-precision=2]{5.34}$ \si{\giga\nothing}} & {\boldmath$\num{398069}$} \\ 
        \hline
        
    \end{tabular}
\end{table}

\section{Conclusion}
\label{sec:conclusion}
In this paper, we have proposed the \acf{ourmodel} model. \ac{ourmodel} compresses \aclp{hsi} using a transformer-based autoencoder to leverage long-range spectral dependencies.
To enhance training efficiency, we have exploited only a small, randomized subset of the available training set in each epoch.
Experimental results have shown the efficiency of our proposed model in both training time and computational complexity, while also surpassing state-of-the-art \aclp{cae} in terms of reconstruction quality for fixed \aclp{cr}.
As future work, we plan to extend our \ac{ourmodel} model to achieve efficient spatio-spectral compression and explore a strategy for selecting informative training samples.

\bibliographystyle{IEEEtran}
\bibliography{refs.bib}

\end{document}